\documentclass[3p,times,10pt,authoryear]{elsarticle}

\usepackage{amssymb}
\usepackage{amsmath}
\usepackage{booktabs}
\usepackage{graphicx}
\usepackage{array}
\usepackage{tabularx}
\usepackage{adjustbox}
\usepackage{threeparttable}
\usepackage{multirow}
\usepackage{enumitem}
\usepackage{url}
\usepackage{placeins}
\usepackage{needspace}
\usepackage{caption}
\usepackage{tikz}
\usepackage{natbib}
\usetikzlibrary{arrows.meta,positioning}
\newcolumntype{L}[1]{>{\raggedright\arraybackslash}p{#1}}

\setlist[itemize]{itemsep=0pt, parsep=0pt, topsep=2pt, leftmargin=1.5em}
\setlist[enumerate]{itemsep=0pt, parsep=0pt, topsep=2pt, leftmargin=1.5em}
\AtBeginEnvironment{table}{\small}
\AtBeginEnvironment{figure}{\small}

\newcommand{\compacttable}{%
\small
\setlength{\tabcolsep}{3.5pt}%
\renewcommand{\arraystretch}{0.95}%
}

\journal{Journal to be confirmed}

\begin{document}

\begin{frontmatter}

\title{From Structural Equation Modelling to Double Machine Learning: Robustness Analysis for Survey-Based Research}

\author[UniSQ]{Ka Ching Chan}
\author[UniSQ]{Qiana Liu}
\author[UniSQ]{Sanjib Tiwari}
\author[UniSQ]{Ranga Chimhundu}
\affiliation[UniSQ]{organization={School of Business, Law, Humanities and Pathways},
             addressline={University of Southern Queensland},
             city={Springfield},
             postcode={4300},
             state={Queensland},
             country={Australia}}

\begin{abstract}
Structural equation modelling (SEM) is widely used in survey-based business and information systems research to assess latent constructs and theory-driven structural relationships. However, SEM path significance is obtained within a particular model specification and may not show whether findings remain stable under alternative estimation frameworks. This study develops and demonstrates a staged robustness analysis framework that connects SEM, ordinary least squares (OLS) regression, and Double Machine Learning (DML). SEM is first used to refine the measurement structure and estimate the robustness-baseline SEM model, in which the full theory-specified structural path system is retained for downstream robustness analysis before final structural path evaluation. OLS regression is then applied to SEM-derived construct scores as a transparent regression benchmark. Finally, DML-style residualisation is used to examine whether each tested focal relationship remains stable after flexible machine-learning-based adjustment for observed controls. Learner-sensitivity checks compare Random Forest, Gradient Boosting, and Support Vector Machine learners, and selected reverse-direction diagnostics are used to examine directional sensitivity. The framework is demonstrated using a FinTech Digital Customer Intimacy survey model. The findings identify which relationships are stable across SEM, OLS, and DML-style checks, and which require more cautious interpretation. A reproducible Google Colab workbook and generated result files are publicly available, providing a reusable template that researchers and students can adapt to other survey-based latent-construct studies. The paper contributes a practical robustness workflow and interpretation guide for survey-based researchers seeking to complement SEM with conventional and machine-learning-based robustness checks.
\end{abstract}

\begin{keyword}
Structural Equation Modelling \sep Double Machine Learning \sep OLS \sep Robustness Analysis \sep Survey Research \sep Information Systems \sep FinTech
\end{keyword}

\end{frontmatter}

\section{Introduction}

Survey-based research frequently relies on latent constructs such as trust, satisfaction, attitude, intention, perceived quality, and customer intimacy. These constructs are usually measured through multiple survey items rather than observed directly. Structural equation modelling (SEM) is therefore well-suited to this type of research because it enables researchers to assess the measurement model and estimate theory-driven structural relationships within a single framework. However, a significant SEM path coefficient remains a result within a particular model specification. Researchers may therefore benefit from supplementary robustness analyses that examine whether key relationships remain stable under alternative score representations and estimation assumptions before final model decisions are made.

This study develops a staged SEM--OLS--DML robustness framework for survey-based latent-construct research. The framework is organised around two analytical layers. The first layer generates the empirical evidence: the measurement model is refined, the full theory-specified structural model is re-estimated as the robustness-baseline SEM model, construct scores are constructed, and OLS and DML-style robustness checks are run for all SEM paths before final structural path evaluation; SEM factor scores and mean composite scores are constructed, and OLS and DML-style robustness checks are run for all tested SEM paths. The second layer compares and interprets the generated outputs across methods, score representations, nuisance learners, and, where theoretically useful, selected reverse-direction diagnostic checks. This two-layer design separates the production of robustness evidence from the interpretation of convergence, divergence, and sensitivity.

The paper is motivated by an emerging methodological gap. Recent studies have introduced Double Machine Learning (DML) to information systems research \citep{shi2025dmlguide}, combined PLS-SEM with machine-learning algorithms for predictive and exploratory purposes \citep{richter2024plsml}, and applied DML directly in FinTech-related empirical settings \citep{wu2024fintechgreen}. These studies show that machine learning and DML are increasingly relevant to empirical research. Less attention, however, has been given to using DML as a post-SEM, path-by-path robustness tool for survey-based latent-construct models. This paper addresses that gap by positioning SEM as the primary measurement and theory-testing framework, OLS as a transparent regression benchmark, and DML as a flexible-control robustness check for SEM-implied relationships.

The study is guided by the following research questions:

\begin{itemize}
\item \textbf{RQ1:} How can SEM, OLS, and DML-style analysis be integrated into a staged robustness framework for survey-based latent-construct research?

\item \textbf{RQ2:} How can a robustness-baseline SEM model be translated into path-level OLS and DML-style specifications before final structural path evaluation?

\item \textbf{RQ3:} How do OLS and DML-style estimates complement SEM when evaluating the robustness of initially tested structural paths?

\item \textbf{RQ4:} How can researchers interpret convergence, divergence, score sensitivity, learner sensitivity, and selected reverse-direction diagnostics within the proposed robustness framework?

\item \textbf{RQ5:} In the FinTech DCI demonstration case, which initially tested structural paths remain robust across SEM, OLS, and DML-style checks, and which paths require further theory, measurement, or directional-diagnostic discussion?
\end{itemize}

The paper makes three contributions. First, it provides a practical workflow for translating SEM-implied structural paths into OLS and DML-style robustness checks without treating machine learning as a replacement for SEM. Second, it clarifies how score-representation sensitivity and learner sensitivity can be interpreted in survey-based latent-construct research. Third, it demonstrates the workflow using a FinTech Digital Customer Intimacy model in which robustness evidence is generated before final structural paths are evaluated and decisions made, and will provide a Google Colab workbook and generated CSV outputs through a Zenodo archive, allowing researchers and students to adapt the workflow to other survey-based SEM applications.

\section{Literature Background and Methodological Positioning}

\subsection{SEM and latent-construct survey research}

Survey-based studies commonly use multi-item instruments to measure constructs that cannot be directly observed. In this setting, SEM is not merely a regression model with named variables. Its distinctive value lies in modelling latent constructs from multiple observed indicators, assessing measurement quality, and estimating a theory-specified system of structural paths \citep{bollen1989sem,anderson1988sem,mackenzie2011construct}. This measurement role is particularly important in behavioural, business, and information systems research, where construct validity is central to credible theory testing. The two-step logic of first assessing the measurement model and then estimating the structural model provides the foundation for using SEM as the primary modelling stage rather than as a preliminary score-generation tool \citep{anderson1988sem}.

Regression-based robustness checks can still play a useful supplementary role after SEM. SEM and regression have long been discussed as complementary rather than mutually exclusive approaches in applied information systems research \citep{gefen2000semregression}. In the present framework, OLS is used as a transparent benchmark that asks whether SEM-implied relationships remain visible when latent constructs are represented as respondent-level construct scores. This benchmark does not replace SEM; instead, it provides a simpler score-based comparison for examining whether the structural pattern is visible under conventional linear regression assumptions.

\subsection{DML, SEM--ML integration, and adjacent applied research}

DML has become influential because it allows a low-dimensional focal relationship to be estimated while flexible machine-learning methods approximate nuisance functions for observed controls and co-predictors \citep{chernozhukov2018dml}. The DML software framework further supports reproducible implementation of this logic in Python \citep{bach2022doubleml}. Recent information systems work has introduced DML as a flexible semiparametric method for empirical model specification and causal-inference-oriented research designs \citep{shi2025dmlguide}. In applied FinTech and sustainability research, DML is often used directly as the main empirical strategy. For example, \citet{wu2024fintechgreen} investigates the relationship between FinTech development and inclusive green growth using city-level panel data under flexible control adjustment.

A related but distinct stream combines SEM or PLS-SEM with machine learning. For instance, \citet{richter2024plsml}, discuss how PLS-SEM and selected machine-learning algorithms can be combined to improve prediction, identify nonlinearities and interactions, and support theoretical insight in business research. Such work shows the increasing value of SEM--ML integration, but its emphasis is often on predictive, exploratory, or causal-predictive approaches. The present study is narrower and more diagnostic. It asks whether theory-specified SEM paths remain directionally and statistically stable when re-examined using OLS and DML-style robustness checks.

\subsection{Why DML is applied after SEM and path by path}

The distinction is important because applying DML directly to raw survey items or simple composite scores may underuse one of SEM's central strengths: the ability to model measurement structure before testing structural relationships. Therefore, this study first uses SEM to validate the latent-variable measurement structure and identify the retained theoretical path system. OLS and DML are then applied on a per-path basis as supplementary robustness analyses. Each selected SEM path is re-examined as a focal relationship, with other theoretically relevant constructs and observed controls treated as adjustment variables.

This use of DML differs from typical applied DML studies in both purpose and implementation. In many applications, the main objective is to estimate one focal treatment--outcome relationship while flexibly adjusting for controls. In contrast, this study uses DML as a robustness diagnostic for a latent-construct path system. SEM provides the measurement and theoretical map, OLS provides a transparent regression benchmark, and DML provides a flexible-control robustness check for individual SEM-implied paths.

\subsection{Causal caution, endogeneity, and directional interpretation}

Although DML is often discussed within the causal-inference literature, the present study uses it as a robustness analysis rather than as definitive causal identification. The survey data are observational, and the structural paths estimated in SEM, OLS, and DML are therefore interpreted as theoretically grounded associations unless stronger identification assumptions are justified. This cautious interpretation is consistent with Pearl's account of causal models and inference \citep{pearl2009causality} and with Pearl's hierarchy, which distinguishes associational, interventional, and counterfactual claims \citep{bareinboim2022pearlhierarchy}. Accordingly, the proposed SEM--OLS--DML workflow strengthens robustness assessment, but it does not by itself establish intervention-level or counterfactual causal claims.

The same caution applies to exogeneity, endogeneity, and directionality. In SEM, exogenous and endogenous constructs are defined by their positions in the specified structural model. This modelling distinction should not be confused with econometric endogeneity, which refers to potential bias arising from omitted variables, simultaneity, reverse causality, measurement error, or related sources. DML can improve robustness by flexibly adjusting for observed covariates, but it does not eliminate unobserved confounding or prove causal direction in cross-sectional survey data. Throughout this study, directional language refers to the theoretically specified and statistically estimated direction of association among constructs, rather than definitive causal direction. Positive coefficients indicate a direct relationship between the predictor and outcome variables, while negative coefficients indicate an inverse relationship. For theoretically sensitive paths, reverse-direction OLS/DML checks may therefore be used as diagnostic evidence of directional robustness or possible reciprocal association, rather than as proof of bidirectional causality.

\section{Preliminaries and Notation}
\label{sec:preliminaries-notation}

This section introduces the notation used to connect the SEM, OLS, and DML-style robustness stages. The purpose is not to derive a new estimator, but to make explicit how survey indicators, latent construct scores, structural paths, and robustness estimates are represented consistently across the three stages.

\subsection{Observed survey data}
\label{subsec:observed-survey-data}

Let $N$ and $P$ denote the number of respondents and observed survey variables, respectively. For respondent $i$, the full observed response vector is denoted by $\mathbf{Z}_i$:

\begin{equation}
\begin{aligned}
\mathbf{Z}_i
&=
\begin{bmatrix}
Z_{i1} & Z_{i2} & \cdots & Z_{iP}
\end{bmatrix}^{\prime},
\quad
i = 1,\ldots,N.
\end{aligned}
\label{eq:observed-response-vector}
\end{equation}

Stacking all respondent-level vectors gives the observed data matrix:

\begin{equation}
\begin{aligned}
\mathbf{Z}
&=
\begin{bmatrix}
\mathbf{Z}_1^{\prime} \\
\mathbf{Z}_2^{\prime} \\
\vdots \\
\mathbf{Z}_N^{\prime}
\end{bmatrix}
\in
\mathbb{R}^{N \times P}.
\end{aligned}
\label{eq:observed-data-matrix}
\end{equation}

Let $\mathcal{C}$ denote the set of latent constructs and let $\mathcal{I}_c$ denote the set of observed indicators assigned to construct $c \in \mathcal{C}$. If $\mathcal{I}_c = \{j_1,j_2,\ldots,j_{J_c}\}$, then the observed indicator vector for construct $c$ and respondent $i$ is:

\begin{equation}
\begin{aligned}
\mathbf{z}_{c,i}
&=
\begin{bmatrix}
Z_{ij_1} & Z_{ij_2} & \cdots & Z_{ij_{J_c}}
\end{bmatrix}^{\prime},
\quad
j_1,\ldots,j_{J_c} \in \mathcal{I}_c.
\end{aligned}
\label{eq:construct-indicator-vector}
\end{equation}

\subsection{First-order and second-order latent constructs}
\label{subsec:first-second-order-constructs}

The SEM measurement model includes both first-order and second-order latent constructs. For a first-order construct $c$, let $\mathbf{z}_{c,i}$ denote the vector of observed indicators for respondent $i$, and let $\omega_{c,i}$ denote the corresponding latent construct. The first-order measurement model can be written as:

\begin{equation}
\begin{aligned}
\mathbf{z}_{c,i}
&=
\boldsymbol{\lambda}_{c}
\omega_{c,i}
+
\mathbf{e}_{c,i},
\end{aligned}
\label{eq:first-order-measurement}
\end{equation}

where $\boldsymbol{\lambda}_{c}$ is the vector of factor loadings and $\mathbf{e}_{c,i}$ is the vector of measurement errors.

For a second-order construct $h$, the construct is inferred from multiple first-order latent dimensions. Let $\boldsymbol{\omega}_{h,i}^{(1)}$ denote the vector of first-order latent dimensions associated with second-order construct $h$, and let $\omega_{h,i}^{(2)}$ denote the second-order latent construct. The second-order measurement relationship can be represented as:

\begin{equation}
\begin{aligned}
\boldsymbol{\omega}_{h,i}^{(1)}
&=
\boldsymbol{\lambda}_{h}^{(2)}
\omega_{h,i}^{(2)}
+
\boldsymbol{u}_{h,i},
\end{aligned}
\label{eq:second-order-measurement}
\end{equation}

where $\boldsymbol{\lambda}_{h}^{(2)}$ contains the second-order loadings and $\boldsymbol{u}_{h,i}$ captures first-order dimension disturbances not explained by the second-order construct.

In this study, Perceived Quality (PQ) and User Experience (UX) are treated as second-order constructs. PQ is inferred from system, information, and service quality. UX is inferred from utilitarian, hedonic, personalised, and security experiences. Other constructs are treated as first-order latent constructs measured directly by their observed survey indicators.

\subsection{SEM measurement and structural notation}
\label{subsec:sem-measurement-structural-notation}

In the SEM stage, the measurement model defines latent constructs, while the structural model specifies their relationships. Let $\boldsymbol{\xi}_i$ denote the vector of structurally exogenous latent constructs for respondent $i$, and let $\boldsymbol{\eta}_i$ denote the vector of structurally endogenous latent constructs. The combined latent construct vector is:

\begin{equation}
\begin{aligned}
\boldsymbol{\omega}_i
&=
\begin{bmatrix}
\boldsymbol{\xi}_i^{\prime} &
\boldsymbol{\eta}_i^{\prime}
\end{bmatrix}^{\prime}.
\end{aligned}
\label{eq:combined-latent-vector}
\end{equation}

A compact measurement model can be written as:

\begin{equation}
\begin{aligned}
\mathbf{Z}_i
&=
\boldsymbol{\Lambda}
\boldsymbol{\omega}_i
+
\mathbf{e}_i,
\end{aligned}
\label{eq:compact-measurement-model}
\end{equation}

where $\boldsymbol{\Lambda}$ is the factor-loading matrix and $\mathbf{e}_i$ is the vector of measurement errors.

The structural component of the SEM can be written as:

\begin{equation}
\begin{aligned}
\boldsymbol{\eta}_i
&=
\mathbf{B}
\boldsymbol{\eta}_i
+
\boldsymbol{\Gamma}
\boldsymbol{\xi}_i
+
\boldsymbol{\zeta}_i,
\end{aligned}
\label{eq:sem-structural-model}
\end{equation}

where $\mathbf{B}$ contains relationships among endogenous latent constructs, $\boldsymbol{\Gamma}$ contains effects from exogenous to endogenous latent constructs, and $\boldsymbol{\zeta}_i$ is the structural disturbance vector.

The SEM parameter vector is estimated by fitting the model-implied covariance structure to the observed data:

\begin{equation}
\begin{aligned}
\widehat{\boldsymbol{\theta}}
&=
\arg\min_{\boldsymbol{\theta}}
\mathcal{L}_{\mathrm{SEM}}
\left(
\boldsymbol{\theta};
\mathbf{Z},
\mathcal{M},
\mathcal{P}
\right),
\end{aligned}
\label{eq:sem-estimation-objective}
\end{equation}

where $\mathcal{M}$ denotes the measurement specification and $\mathcal{P}$ denotes the structural path specification.

In SEM terminology, structurally exogenous and endogenous constructs are distinguished by their positions in the specified structural model. Exogenous constructs do not receive incoming structural paths, whereas endogenous constructs are explained by one or more other constructs. This distinction is important for SEM specification and interpretation. However, once respondent-level construct scores are extracted, both exogenous and endogenous constructs can be represented within the same construct-score matrix for downstream OLS and DML-style robustness checks.

\subsection{Construct-score representations}
\label{subsec:construct-score-representations}

After estimating the SEM, respondent-level latent construct scores are extracted. Let $\widehat{\boldsymbol{\omega}}_i$ denote the SEM factor-score vector for respondent $i$. A general model-based factor-score representation is:

\begin{equation}
\begin{aligned}
\widehat{\boldsymbol{\omega}}_i
&=
\boldsymbol{\Phi}_{\omega}
\boldsymbol{\Lambda}^{\prime}
\mathbf{\Sigma}
\left(
\widehat{\boldsymbol{\theta}}
\right)^{-1}
\mathbf{Z}_i,
\end{aligned}
\label{eq:sem-factor-score-extraction}
\end{equation}

where $\boldsymbol{\Phi}_{\omega}$ is the latent construct covariance matrix, $\boldsymbol{\Lambda}$ is the estimated loading matrix, and $\mathbf{\Sigma}(\widehat{\boldsymbol{\theta}})$ is the model-implied covariance matrix.

Stacking all extracted SEM factor scores gives:

\begin{equation}
\begin{aligned}
\widehat{\mathbf{F}}^{\mathrm{SEM}}
&=
\begin{bmatrix}
\widehat{\boldsymbol{\omega}}_1^{\prime} \\
\widehat{\boldsymbol{\omega}}_2^{\prime} \\
\vdots \\
\widehat{\boldsymbol{\omega}}_N^{\prime}
\end{bmatrix}
\in
\mathbb{R}^{N \times C},
\end{aligned}
\label{eq:sem-factor-score-matrix}
\end{equation}

where $C$ is the number of construct scores used in the downstream robustness checks.

As a simpler benchmark, mean composite scores are calculated from the retained indicators. For construct $c$ and respondent $i$, the composite score is:

\begin{equation}
\begin{aligned}
\bar{Z}_{c,i}^{\mathrm{COMP}}
&=
\frac{1}{J_c}
\sum_{j \in \mathcal{I}_c}
Z_{ij}.
\end{aligned}
\label{eq:mean-composite-score}
\end{equation}

Stacking the composite scores gives the composite-score matrix:

\begin{equation}
\begin{aligned}
\mathbf{F}^{\mathrm{COMP}}
&=
\begin{bmatrix}
\bar{\mathbf{z}}_1^{\mathrm{COMP}\prime} \\
\bar{\mathbf{z}}_2^{\mathrm{COMP}\prime} \\
\vdots \\
\bar{\mathbf{z}}_N^{\mathrm{COMP}\prime}
\end{bmatrix}
\in
\mathbb{R}^{N \times C}.
\end{aligned}
\label{eq:composite-score-matrix}
\end{equation}

The robustness checks therefore use two alternative construct-score representations:

\begin{equation}
\begin{aligned}
s
&\in
\left\{
\mathrm{SEM},
\mathrm{COMP}
\right\},
\end{aligned}
\label{eq:score-representation-set}
\end{equation}

where $\mathrm{SEM}$ denotes SEM factor scores and $\mathrm{COMP}$ denotes mean composite scores.

SEM factor scores are treated as the primary construct-score representation because they preserve the fitted SEM measurement model, including estimated factor loadings, latent-variable structure, and the distinction between first-order and second-order constructs. Mean composite scores are included as a transparent sensitivity benchmark. Composite scores are widely used in survey-based research because they are simple, interpretable, and easy to reproduce. However, they treat retained indicators as equally weighted and do not explicitly account for factor loadings, measurement error, or higher-order construct structure. Comparing the two score representations is therefore useful for diagnosing measurement-representation sensitivity.
The next section uses this notation to specify how tested SEM paths are translated into OLS and DML-style robustness checks.

\section{Proposed SEM--OLS--DML Robustness Framework}

\subsection{Methodological rationale}

The proposed framework is based on methodological complementarity rather than methodological substitution. SEM remains the primary modelling stage because it validates the latent measurement structure and estimates theory-driven structural relationships. OLS provides a transparent score-based benchmark by asking whether the SEM-implied relationships remain visible when respondent-level construct scores are analysed using conventional regression. DML-style residualisation provides an additional robustness layer by asking whether focal path estimates remain stable when the outcome and focal predictor are flexibly adjusted for observed controls and co-predictors. The three methods therefore answer different but connected questions: SEM asks whether the latent-variable model is theoretically and statistically supported; OLS asks whether the same path pattern is visible under a simpler linear score-based specification; and DML-style checks ask whether the focal relationships remain stable under flexible control-function adjustment.

Operationally, the framework separates the analysis into two layers. The \emph{result-generation layer} estimates the SEM model, constructs alternative construct-score representations, and runs the OLS and DML-style models across the planned scenarios. The \emph{comparison-and-interpretation layer} then compares the generated outputs across methods, score representations, nuisance learners, and selected reverse-direction diagnostics. This separation is useful because the Colab implementation can first export all scenario-level results as CSV files, after which the paper can analyse convergence, divergence, and sensitivity in an integrated way.

\subsection{Workflow overview}

The proposed workflow is organised into eight operational stages within these two analytical layers. Stages 1--7 form the result-generation layer: they refine the measurement model, establish the robustness-baseline SEM model, construct factor and composite scores, define path-level specifications, and run OLS and DML-style checks across score representations and nuisance learners before final structural path evaluation. Stage 8 forms the comparison-and-interpretation layer: it integrates the evidence by comparing direction, statistical support, score-representation sensitivity, learner sensitivity, and selected reverse-direction diagnostics to support final structural model refinement decisions.

The workflow distinguishes measurement refinement from final structural path evaluation. The initial conceptual model specifies both the measurement structure and the theory-driven structural paths among constructs. The measurement part is first evaluated using item diagnostics, factor loadings, reliability, validity, and model-fit evidence. Where justified, indicators or measurement factors are refined so that the constructs used in subsequent analysis are empirically defensible. However, the theory-specified structural paths are not removed or modified immediately after SEM estimation solely on the basis of SEM significance. Instead, after the refined measurement model is obtained, the full theory-specified structural model is re-estimated and treated as the robustness-baseline SEM model. This model provides the path system that is then translated into OLS and DML-style specifications. Final structural model refinement decisions are made only after SEM estimates, OLS benchmarks, DML-style checks, score sensitivity, learner sensitivity, reverse-direction diagnostics, and theoretical considerations are examined together.

Table~\ref{tab:framework_stages} complements the figure by translating the same workflow into operational stages, questions, and outputs that can be followed in empirical applications.

\begin{table}[!t]
\centering
\caption{Operational stages of the SEM--OLS--DML robustness workflow}
\label{tab:framework_stages}
\compacttable
\renewcommand{\arraystretch}{1.2}
\begin{tabular}{L{0.06\linewidth} L{0.2\linewidth} L{0.36\linewidth} L{0.28\linewidth}}
\toprule
\textbf{Stage} & \textbf{Workflow component} & \textbf{Main question} & \textbf{Analytical output} \\
\midrule
1 & Initial conceptual model and item diagnostics & Are the proposed constructs, indicators, and theory-specified structural paths suitable for empirical testing? & Initial conceptual model and diagnostic evidence for item/indicator quality \\

2 & Measurement-model refinement & Do the retained indicators provide an empirically defensible measurement model for the latent constructs? & Refined measurement model, retained indicators, reliability, validity, and fit evidence \\

3 & Robustness-baseline SEM estimation & Using the refined measurement model, what evidence is obtained when the full theory-specified structural model is estimated before final structural path evaluation? & Robustness-baseline SEM model, SEM path estimates, and tested structural path system \\

4 & Construct-score construction & How are the refined latent constructs represented for downstream robustness analysis? & SEM factor scores and mean composite scores \\

5 & SEM-to-regression path translation & How is each tested SEM path translated into path-level OLS and DML-style specifications?
& Outcome construct $Y$, focal predictor $D$, and control vector $\mathbf{X}$ for each tested path \\

6 & OLS robustness benchmark & Do the tested SEM paths remain visible under transparent score-based linear regression? & OLS path-level robustness estimates across construct-score representations \\

7 & DML-style robustness and sensitivity checks & Do the tested SEM paths remain stable after flexible adjustment for SEM co-predictors and observed controls, and are results sensitive to the learner used?
& DML-style estimates across score types and RF, GBM, and SVM learners \\

8 & Integrated robustness interpretation and final structural path evaluation & Where do SEM, OLS, and DML-style evidence converge or diverge, and which paths should be retained, reviewed, or treated as directionally sensitive? & Integrated robustness evidence, selected reverse-direction diagnostics, and final structural model refinement decisions \\

\bottomrule
\multicolumn{4}{p{0.94\linewidth}}{\footnotesize \textit{Note.} The workflow separates measurement refinement from structural path evaluation. Indicators and measurement factors may be refined before the robustness analysis, but theory-specified structural paths are not modified immediately after SEM estimation. Final structural model refinement decisions are made only after SEM, OLS, DML-style results, score sensitivity, learner sensitivity, reverse-direction diagnostics, and theoretical considerations are reviewed together.}
\end{tabular}
\end{table}

\subsection{SEM-to-regression translation}

A SEM structural path can be translated into an OLS or DML-style specification by identifying the outcome construct, the focal predictor, and the local co-predictor/control set. For a SEM equation of the form $Y \sim D + X$, OLS estimates the score-based regression of $Y$ on $D$ and $X$, while DML-style residualisation partials out $X$ from both $Y$ and $D$ before estimating the residual relationship.

A key feature of this translation is that SEM structural equations are translated at the equation level, while the resulting regression coefficients are interpreted at the path level. One regression equation may therefore contain multiple SEM-implied paths. For example, the SEM equation $BI \leftarrow AT + CA$ corresponds to one regression model with $BI$ as the outcome and $AT$ and $CA$ as predictors, but it yields two path-level benchmarks: $AT \rightarrow BI$ and $CA \rightarrow BI$. Similarly, $SA \leftarrow BI + UX + UB$ corresponds to one regression equation but provides three path-level benchmarks. This distinction is important because the robustness analysis follows the SEM structural system while allowing each individual path coefficient to be compared across SEM, OLS, and DML.

\begin{table}[!t]
\centering
\caption{SEM-to-regression translation logic}
\label{tab:sem_to_regression}
\compacttable
\renewcommand{\arraystretch}{1.2}
\begin{tabular}{L{0.29\linewidth} L{0.25\linewidth} L{0.43\linewidth}}
\toprule
\textbf{SEM component} & \textbf{OLS/DML translation} & \textbf{Purpose} \\
\midrule
Latent construct & Factor or composite score & Enables construct-level regression \\
Path $Y \sim D$ & Focal relationship & Defines the path to check \\
Companion predictors & Control set $X$ & Preserves local structural context \\
Robustness-baseline SEM estimation & Benchmark theory-test result & Primary evidence from latent-variable model \\
OLS coefficient & Linear score-based estimate & Transparent robustness check \\
DML-style coefficient & Residualised focal estimate & Robustness after flexible observed-control adjustment \\
\bottomrule
\end{tabular}
\end{table}

\subsection{Path-level robustness notation}
\label{subsec:path-level-robustness-notation}

Let $k = 1,\ldots,K$ index the structural paths from the robustness-baseline SEM model. For each path $k$ and score representation $s \in \{\mathrm{SEM},\mathrm{COMP}\}$, let $Y_{ik}^{(s)}$ denote the outcome construct score, $D_{ik}^{(s)}$ denote the focal predictor construct score, and $\mathbf{X}_{ik}^{(s)}$ denote the control vector for respondent $i$. The control vector may include observed controls and SEM co-predictors from the same structural equation. For example, when a SEM equation contains two predictors, the focal predictor is treated as $D$, while the other predictor is included in $\mathbf{X}$ as a co-predictor control.

The path-level notation is:

\begin{equation}
\begin{aligned}
Y_{ik}^{(s)}
&= \text{outcome construct score}, \\
D_{ik}^{(s)}
&= \text{focal predictor construct score}, \\
\mathbf{X}_{ik}^{(s)}
&= \text{observed controls and co-predictors}.
\end{aligned}
\label{eq:path-level-notation}
\end{equation}

\subsection{OLS robustness benchmark}
\label{subsec:ols-robustness-benchmark}

OLS is included as an intermediate benchmark rather than as a competing method. Because DML estimates can be less transparent to survey researchers, OLS provides a simple regression-based reference point between SEM and the more flexible DML analysis. This benchmark helps distinguish relationships that are sensitive to the move from SEM to construct-score regression from those that are specifically sensitive to machine-learning-based control adjustment.

Following the SEM-to-regression translation described above, OLS is estimated equation by equation using the selected construct-score representation, and the resulting coefficients are interpreted as path-level benchmarks. For each focal SEM path $k$, the corresponding OLS benchmark can be written as:

\begin{equation}
\begin{aligned}
Y_{ik}^{(s)}
&=
\alpha_k^{(s)}
+
\theta_{\mathrm{OLS},k}^{(s)}
D_{ik}^{(s)}
+
\boldsymbol{\gamma}_k^{(s)\prime}
\mathbf{X}_{ik}^{(s)}
+
\varepsilon_{ik}^{(s)}.
\end{aligned}
\label{eq:ols-path-model}
\end{equation}

Here, $Y_{ik}^{(s)}$ is the outcome construct score for respondent $i$ in focal path $k$ under score representation $s$, $D_{ik}^{(s)}$ is the focal predictor construct score, and $\mathbf{X}_{ik}^{(s)}$ contains companion predictors or theoretically relevant controls included in the corresponding SEM structural equation. The coefficient $\theta_{\mathrm{OLS},k}^{(s)}$ is the OLS benchmark estimate for the focal SEM path, and $\varepsilon_{ik}^{(s)}$ is the regression error term.

This stage deliberately uses a linear additive score-based specification. Its value is therefore transparency: OLS provides a familiar benchmark for examining whether the SEM-implied path pattern remains visible under conventional regression assumptions before the more flexible DML checks are conducted.

\subsection{DML-style partialling-out robustness check}
\label{subsec:dml-style-partialling-out}

The DML-style robustness check uses a partially linear residualisation logic. For each tested path $k$, the outcome and focal predictor are modelled as functions of the observed controls and co-predictors:

\begin{equation}
\begin{aligned}
Y_{ik}^{(s)}
&=
\theta_{\mathrm{DML},k}^{(s)}
D_{ik}^{(s)}
+
g_k^{(s)}
\left(
\mathbf{X}_{ik}^{(s)}
\right)
+
\varepsilon_{ik}^{(s)}, \\
D_{ik}^{(s)}
&=
m_k^{(s)}
\left(
\mathbf{X}_{ik}^{(s)}
\right)
+
\nu_{ik}^{(s)}.
\end{aligned}
\label{eq:dml-partially-linear-model}
\end{equation}

The nuisance functions $g_k^{(s)}(\cdot)$ and $m_k^{(s)}(\cdot)$ are estimated using machine-learning models. In contrast with OLS, the DML-style stage does not require the control functions for the outcome and focal predictor to be linear in $\mathbf{X}_{ik}^{(s)}$. The focal path remains partially linear, but the adjustment for observed controls and co-predictors can be estimated flexibly. This makes the DML-style check useful for diagnosing whether the OLS robustness result depends on a simple linear control-function assumption. The nuisance functions are represented as:

\begin{equation}
\begin{aligned}
\widehat{g}_k^{(s)}
\left(
\mathbf{X}_{ik}^{(s)}
\right)
&\approx
\mathbb{E}
\left[
Y_{ik}^{(s)}
\mid
\mathbf{X}_{ik}^{(s)}
\right], \\
\widehat{m}_k^{(s)}
\left(
\mathbf{X}_{ik}^{(s)}
\right)
&\approx
\mathbb{E}
\left[
D_{ik}^{(s)}
\mid
\mathbf{X}_{ik}^{(s)}
\right].
\end{aligned}
\label{eq:dml-nuisance-functions}
\end{equation}

The residualised outcome and residualised focal predictor are then:

\begin{equation}
\begin{aligned}
\widetilde{Y}_{ik}^{(s)}
&=
Y_{ik}^{(s)}
-
\widehat{g}_k^{(s)}
\left(
\mathbf{X}_{ik}^{(s)}
\right), \\
\widetilde{D}_{ik}^{(s)}
&=
D_{ik}^{(s)}
-
\widehat{m}_k^{(s)}
\left(
\mathbf{X}_{ik}^{(s)}
\right).
\end{aligned}
\label{eq:dml-residuals}
\end{equation}

The DML-style estimate is obtained by regressing the residualised outcome on the residualised focal predictor:

\begin{equation}
\begin{aligned}
\widehat{\theta}_{\mathrm{DML},k}^{(s)}
&=
\left(
\widetilde{\mathbf{D}}_{k}^{(s)\prime}
\widetilde{\mathbf{D}}_{k}^{(s)}
\right)^{-1}
\widetilde{\mathbf{D}}_{k}^{(s)\prime}
\widetilde{\mathbf{Y}}_{k}^{(s)}.
\end{aligned}
\label{eq:dml-residual-regression}
\end{equation}

In the implementation, cross-fitting is used to reduce overfitting in the nuisance-function estimation. Let $\mathcal{I}_q$ denote fold $q$ and let $\mathcal{I}_{-q}$ denote the training observations outside fold $q$. For learner $\ell$ and observations $i \in \mathcal{I}_q$, the cross-fitted residuals are:

\begin{equation}
\begin{aligned}
\widetilde{Y}_{ik,\ell}^{(s)}
&=
Y_{ik}^{(s)}
-
\widehat{g}_{k,\ell,-q}^{(s)}
\left(
\mathbf{X}_{ik}^{(s)}
\right),
\quad
i \in \mathcal{I}_q, \\
\widetilde{D}_{ik,\ell}^{(s)}
&=
D_{ik}^{(s)}
-
\widehat{m}_{k,\ell,-q}^{(s)}
\left(
\mathbf{X}_{ik}^{(s)}
\right),
\quad
i \in \mathcal{I}_q.
\end{aligned}
\label{eq:cross-fitted-residuals}
\end{equation}

Here, $\ell$ indexes the machine-learning learner used to estimate the nuisance functions, and $-q$ indicates that the nuisance model is trained on observations outside fold $q$.

\subsection{Machine-learning nuisance learners}
\label{subsec:ml-nuisance-learners}

To examine whether the DML-style conclusions depend on one particular nuisance learner, the residualisation step is repeated using multiple learners:

\begin{equation}
\begin{aligned}
\mathcal{L}
&=
\left\{
\mathrm{RF},
\mathrm{GBM},
\mathrm{SVM}
\right\}.
\end{aligned}
\label{eq:ml-learner-set}
\end{equation}

For each tested path $k$, score representation $s$, and learner $\ell \in \mathcal{L}$, the learner-specific nuisance functions are:

\begin{equation}
\begin{aligned}
\widehat{g}_{k,\ell}^{(s)}
\left(
\mathbf{X}_{ik}^{(s)}
\right)
&\approx
\mathbb{E}
\left[
Y_{ik}^{(s)}
\mid
\mathbf{X}_{ik}^{(s)}
\right], \\
\widehat{m}_{k,\ell}^{(s)}
\left(
\mathbf{X}_{ik}^{(s)}
\right)
&\approx
\mathbb{E}
\left[
D_{ik}^{(s)}
\mid
\mathbf{X}_{ik}^{(s)}
\right].
\end{aligned}
\label{eq:learner-specific-nuisance-functions}
\end{equation}

Random Forest represents the nuisance prediction as an average of regression trees:

\begin{equation}
\begin{aligned}
\widehat{f}_{\mathrm{RF}}
\left(
\mathbf{X}
\right)
&=
\frac{1}{B}
\sum_{b=1}^{B}
T_b
\left(
\mathbf{X}
\right),
\end{aligned}
\label{eq:rf-prediction}
\end{equation}

where $T_b(\cdot)$ is the prediction from tree $b$ and $B$ is the number of trees.

Gradient Boosting Machine represents the nuisance prediction as an additive sequence of weak learners:

\begin{equation}
\begin{aligned}
\widehat{f}_{\mathrm{GBM}}
\left(
\mathbf{X}
\right)
&=
f_0
\left(
\mathbf{X}
\right)
+
\sum_{m=1}^{M}
\nu
h_m
\left(
\mathbf{X}
\right),
\end{aligned}
\label{eq:gbm-prediction}
\end{equation}

where $f_0(\cdot)$ is the initial prediction function, $h_m(\cdot)$ is the weak learner added at iteration $m$, $M$ is the number of boosting iterations, and $\nu$ is the learning rate.

Support Vector Machine regression can be written as:

\begin{equation}
\begin{aligned}
\widehat{f}_{\mathrm{SVM}}
\left(
\mathbf{X}
\right)
&=
\sum_{r \in \mathcal{SV}}
\left(
\alpha_r
-
\alpha_r^{*}
\right)
K
\left(
\mathbf{X}_r,
\mathbf{X}
\right)
+
b_0,
\end{aligned}
\label{eq:svm-prediction}
\end{equation}

where $\mathcal{SV}$ is the set of support vectors, $K(\cdot,\cdot)$ is the kernel function, $\alpha_r$ and $\alpha_r^{*}$ are support-vector coefficients, and $b_0$ is the intercept term.

These learners are not used to redefine the SEM model, serving solely to estimate nuisance functions within the DML-style residualisation step. Learner-sensitivity analysis examines whether the estimated path direction and statistical support remain stable when the nuisance functions are estimated using RF, GBM, and SVM.

\subsection{Directional robustness and reverse-direction diagnostics}
\label{subsec:directional-robustness-reverse-checks}

For most tested SEM paths, the robustness checks are conducted in the SEM-specified direction. However, selected theoretically sensitive paths may also be examined in the reverse direction as a diagnostic extension. This is useful when a relationship could plausibly be reciprocal, when the SEM sign is unexpected, or when score-representation sensitivity suggests that the path may require cautious interpretation.

For a tested SEM path $D \rightarrow Y$, the same-direction DML-style check treats $Y$ as the outcome and $D$ as the focal predictor. The reverse-direction diagnostic swaps the focal roles:

\begin{equation}
\begin{aligned}
D_{ik}^{(s)}
&=
\phi_{\mathrm{DML},k}^{(s)}Y_{ik}^{(s)}
+
h_k^{(s)}
\left(
\mathbf{X}_{ik}^{(s)}
\right)
+
\upsilon_{ik}^{(s)}.
\end{aligned}
\label{eq:reverse-direction-diagnostic}
\end{equation}

The reverse-direction check is not interpreted as proof of bidirectional causality. Instead, it is a diagnostic test of directional robustness. If the SEM-specified direction is supported but the reverse direction is weak, the original direction has stronger robustness support. If both directions are supported, the relationship may reflect reciprocal association, simultaneity, or shared antecedents. If the reverse direction is stronger than the SEM-specified direction, the original theoretical specification should be interpreted cautiously and may require further longitudinal or experimental validation.

\subsection{Robustness interpretation criteria}
\label{subsec:robustness-interpretation-criteria}

The robustness checks focus on direction, statistical support, score sensitivity, and learner sensitivity. Directional stability across SEM, OLS, and DML-style estimates is defined as:

\begin{equation}
\begin{aligned}
\mathrm{sign}
\left(
\widehat{\beta}_{\mathrm{SEM},k}
\right)
&=
\mathrm{sign}
\left(
\widehat{\theta}_{\mathrm{OLS},k}^{(s)}
\right)
=
\mathrm{sign}
\left(
\widehat{\theta}_{\mathrm{DML},k}^{(s)}
\right).
\end{aligned}
\label{eq:directional-stability}
\end{equation}

Statistical stability requires that the path remains statistically supported across the robustness checks:

\begin{equation}
\begin{aligned}
p_{\mathrm{OLS},k}^{(s)}
&<
\alpha,
\quad
p_{\mathrm{DML},k}^{(s)}
<
\alpha,
\end{aligned}
\label{eq:statistical-stability}
\end{equation}

where $\alpha$ is the chosen significance threshold.

Score sensitivity is assessed by comparing the SEM factor-score and mean composite-score results:

\begin{equation}
\begin{aligned}
\mathrm{sign}
\left(
\widehat{\theta}_{\mathrm{DML},k}^{\mathrm{SEM}}
\right)
&=
\mathrm{sign}
\left(
\widehat{\theta}_{\mathrm{DML},k}^{\mathrm{COMP}}
\right).
\end{aligned}
\label{eq:score-sensitivity-direction}
\end{equation}

If the signs remain stable but statistical support differs across score representations, the path is interpreted as measurement-representation sensitive.

Learner sensitivity is assessed by comparing GBM, SVM, and RF:

\begin{equation}
\begin{aligned}
\mathrm{sign}
\left(
\widehat{\theta}_{\mathrm{DML},k,\mathrm{GBM}}^{(s)}
\right)
&=
\mathrm{sign}
\left(
\widehat{\theta}_{\mathrm{DML},k,\mathrm{SVM}}^{(s)}
\right)
=
\mathrm{sign}
\left(
\widehat{\theta}_{\mathrm{DML},k,\mathrm{RF}}^{(s)}
\right).
\end{aligned}
\label{eq:learner-directional-stability}
\end{equation}

A path is classified as strongly robust when it is directionally and statistically stable across SEM, OLS, and DML-style checks. A path is classified as partially robust when the direction is stable but statistical support weakens under one or more robustness specifications. A path is interpreted cautiously when its sign changes, or when its statistical support depends strongly on the construct-score representation or nuisance learner.

Before applying the workflow to the DCI demonstration case, it is necessary to clarify how convergence and divergence across SEM, OLS, and DML-style checks are interpreted. The three approaches do not estimate identical models: SEM estimates the latent-variable path system, OLS translates the tested SEM equations from the robustness-baseline SEM model into transparent construct-score regressions, and DML-style analysis re-examines each focal path under flexible control-function adjustment. Therefore, robustness is assessed through directional consistency, statistical support, score sensitivity, learner sensitivity, and, for selected sensitive paths, reverse-direction diagnostic evidence. Table~\ref{tab:robustness_interpretation_rules} summarises the interpretation rules used to classify each path in the empirical demonstration.

\begin{table}[!t]
\centering
\caption{Robustness interpretation rules for SEM--OLS--DML comparison}
\label{tab:robustness_interpretation_rules}
\compacttable
\renewcommand{\arraystretch}{1.3}
\begin{tabular}{L{0.25\linewidth} L{0.33\linewidth} L{0.15\linewidth} L{0.20\linewidth}}
\toprule
\textbf{Evidence pattern} & \textbf{Interpretation logic} & \textbf{Robustness classification} & \textbf{Reporting implication} \\
\midrule

SEM, OLS, and DML-style estimates have the same direction and comparable statistical support across score types and learners
&
The relationship remains visible across the latent-variable model, transparent regression benchmark, and flexible-control adjustment
&
Strongly robust
&
Report as a stable SEM-implied relationship \\

SEM and OLS are directionally consistent, but DML-style support weakens or becomes insignificant
&
The relationship survives construct-score regression but is sensitive to flexible control-function adjustment
&
Partially robust / DML-sensitive
&
Interpret cautiously and discuss possible nonlinear control or specification sensitivity \\

SEM is supported, but OLS and DML-style checks are weak or inconsistent
&
The relationship may be more dependent on the SEM system, latent-variable specification, or measurement structure
&
SEM-specific / score-sensitive
&
Report as theoretically relevant but empirically sensitive outside SEM \\

Direction is stable, but significance differs between SEM factor scores and mean composite scores
&
The result depends partly on how latent constructs are represented for downstream robustness checks
&
Score-sensitive
&
Compare factor-score and composite-score results explicitly \\

DML-style estimates differ substantially across RF, GB, and SVM learners
&
The robustness conclusion depends on the nuisance-function learner used in DML-style residualisation
&
Learner-sensitive
&
Avoid strong claims; report learner-level results transparently \\

SEM, OLS, or DML-style estimates show opposite directions across methods or score types
&
The estimated relationship is unstable and may reflect suppression, multicollinearity, measurement issues, or model-specification sensitivity
&
Directionally unstable
&
Interpret with substantial caution and revisit theory or measurement \\

Reverse-direction diagnostics show support only for the SEM-specified direction
&
The specified direction is more empirically stable than the reverse diagnostic direction
&
Directionally supported
&
Use as diagnostic support, not causal proof \\

Reverse-direction diagnostics show support in both directions
&
The constructs may be reciprocally associated, simultaneously related, or shaped by shared antecedents
&
Possible reciprocal association
&
Discuss as directional diagnostic evidence, not proof of bidirectional causality \\

Reverse-direction diagnostics show stronger support for the reverse direction
&
The SEM-specified direction may require theoretical reconsideration or further validation
&
Directionally sensitive
&
Flag for future testing, preferably with longitudinal or experimental data \\

\bottomrule
\end{tabular}
\end{table}


\section{Empirical Demonstration: FinTech Digital Customer Intimacy}

The framework is demonstrated using a FinTech Digital Customer Intimacy (DCI) survey model, \citep{liu2026dcimodel}. The revised implementation estimates the robustness-baseline SEM model after measurement refinement and before final structural path evaluation. This is an important change from a conventional trim-then-check workflow. In the revised workflow, SEM, OLS, DML-style robustness checks, learner-sensitivity checks, and reverse-direction diagnostics are all run on the initially tested SEM path system. Final path decisions are then left to the end, after all empirical evidence has been considered together.

The tested SEM model includes 17 structural paths. Its structural component is:

\begin{equation}
\begin{aligned}
AT_i &= \beta_{1}CA_i + \beta_{2}PQ_i + \zeta_{1,i}, \\
BI_i &= \beta_{3}AT_i + \beta_{4}CA_i + \zeta_{2,i}, \\
UB_i &= \beta_{5}BI_i + \zeta_{3,i}, \\
UX_i &= \beta_{6}UB_i + \beta_{7}PQ_i + \zeta_{4,i}, \\
SA_i &= \beta_{8}BI_i + \beta_{9}UX_i + \beta_{10}UB_i + \zeta_{5,i}, \\
TR_i &= \beta_{11}UB_i + \beta_{12}SA_i + \zeta_{6,i}, \\
EC_i &= \beta_{13}UB_i + \beta_{14}TR_i + \zeta_{7,i}, \\
DCI_i &= \beta_{15}EC_i + \beta_{16}TR_i + \beta_{17}CA_i + \zeta_{8,i}.
\end{aligned}
\label{eq:tested_sem_structural_model}
\end{equation}

Table~\ref{tab:constructs_indicators} summarises the retained indicators and dimensions used to estimate the tested SEM model. The DCI construct uses DCI1, DCI23, and DCI5, where DCI23 is the averaged indicator formed from DCI2 and DCI3.

\begin{table}[!t]
\centering
\caption{Demonstration case constructs and retained indicators in the tested SEM model}
\label{tab:constructs_indicators}
\compacttable
{\renewcommand{\arraystretch}{1.05}
\begin{tabular}{L{0.12\linewidth} L{0.30\linewidth} L{0.38\linewidth} L{0.12\linewidth}}
\toprule
\textbf{Code} & \textbf{Construct} & \textbf{Retained indicators / dimensions} & \textbf{Alpha} \\
\midrule
PQ\_SY & Perceived Quality -- System Quality & PQ\_SY1, PQ\_SY2, PQ\_SY3 & 0.778 \\
PQ\_IQ & Perceived Quality -- Information Quality & PQ\_IQ1, PQ\_IQ2, PQ\_IQ3 & 0.784 \\
PQ\_SQ & Perceived Quality -- Service Quality & PQ\_SQ1, PQ\_SQ2, PQ\_SQ3 & 0.859 \\
PQ & Perceived Quality & PQ\_SY, PQ\_IQ, PQ\_SQ & Second-order \\
CA & Customer Awareness & CA1, CA2, CA3 & 0.744 \\
AT & Attitude & AT1, AT2 & 0.803 \\
BI & Behavioural Intention & BI1, BI2 & 0.828 \\
UB & Usage Behaviour & UB\_UF, UB\_UV2 & 0.409 \\
UX\_PE & User Experience -- Pragmatic Experience & UX\_PE1, UX\_PE2 & 0.816 \\
UX\_SE & User Experience -- Social Experience & UX\_SE1, UX\_SE2 & 0.829 \\
UX & User Experience & UX\_PE, UX\_SE & Second-order \\
SA & Satisfaction & SA1, SA2 & 0.887 \\
TR & Trust & TR1, TR2 & 0.809 \\
EC & Emotional Connection & EC1, EC2, EC3 & 0.826 \\
DCI & Digital Customer Intimacy & DCI1, DCI23, DCI5 & 0.872 \\
\bottomrule
\end{tabular}
}
\vspace{1mm}
\begin{minipage}{0.96\linewidth}
\footnotesize \textit{Note.} DCI23 denotes the averaged indicator formed from DCI2 and DCI3. Alpha is reported for first-order retained constructs only; second-order constructs are represented by their first-order dimensions.
\end{minipage}
\end{table}

\subsection{Robustness-baseline SEM specification and fit}
\label{subsec:tested-sem-specification-fit}

The robustness-baseline DCI SEM model provides the common structural path system for all downstream robustness checks. Table~\ref{tab:sem_fit_comparison} reports the fit overview for this model. The model has acceptable approximate fit for demonstration purposes, with CFI above 0.90, RMSEA below 0.08, and $\chi^2/df$ below 3. The purpose of the demonstration is not to claim that the tested model is the final substantive model, but to show how empirical evidence can be generated before making final structural model refinement decisions.

\begin{table}[!t]
\centering
\caption{Tested SEM model fit overview}
\label{tab:sem_fit_comparison}
\compacttable
\begin{tabular}{L{0.25\linewidth} L{0.08\linewidth} L{0.12\linewidth}L{0.08\linewidth}L{0.08\linewidth}L{0.08\linewidth}L{0.08\linewidth}L{0.08\linewidth}}
\toprule
\textbf{Model} & \textbf{DoF} & \textbf{$\chi^2$} & \textbf{$\chi^2$/df} & \textbf{CFI} & \textbf{TLI} & \textbf{RMSEA} & \textbf{BIC} \\
\midrule
Tested SEM model & 441 & 1166.421 & 2.645 & 0.907 & 0.895 & 0.063 & 517.942 \\
\bottomrule
\end{tabular}
\vspace{1mm}
\begin{minipage}{0.96\linewidth}
\footnotesize \textit{Note.} The robustness-baseline SEM model is the full theory-driven structural specification estimated after measurement refinement and used to generate SEM, OLS, DML, learner-sensitivity, and reverse-diagnostic evidence before final structural model refinement decisions are made.
\end{minipage}
\end{table}

\begin{figure}[!t]
\centering
\includegraphics[width=1.0\linewidth, trim=4.5cm 7cm 4.5cm 5cm, clip]{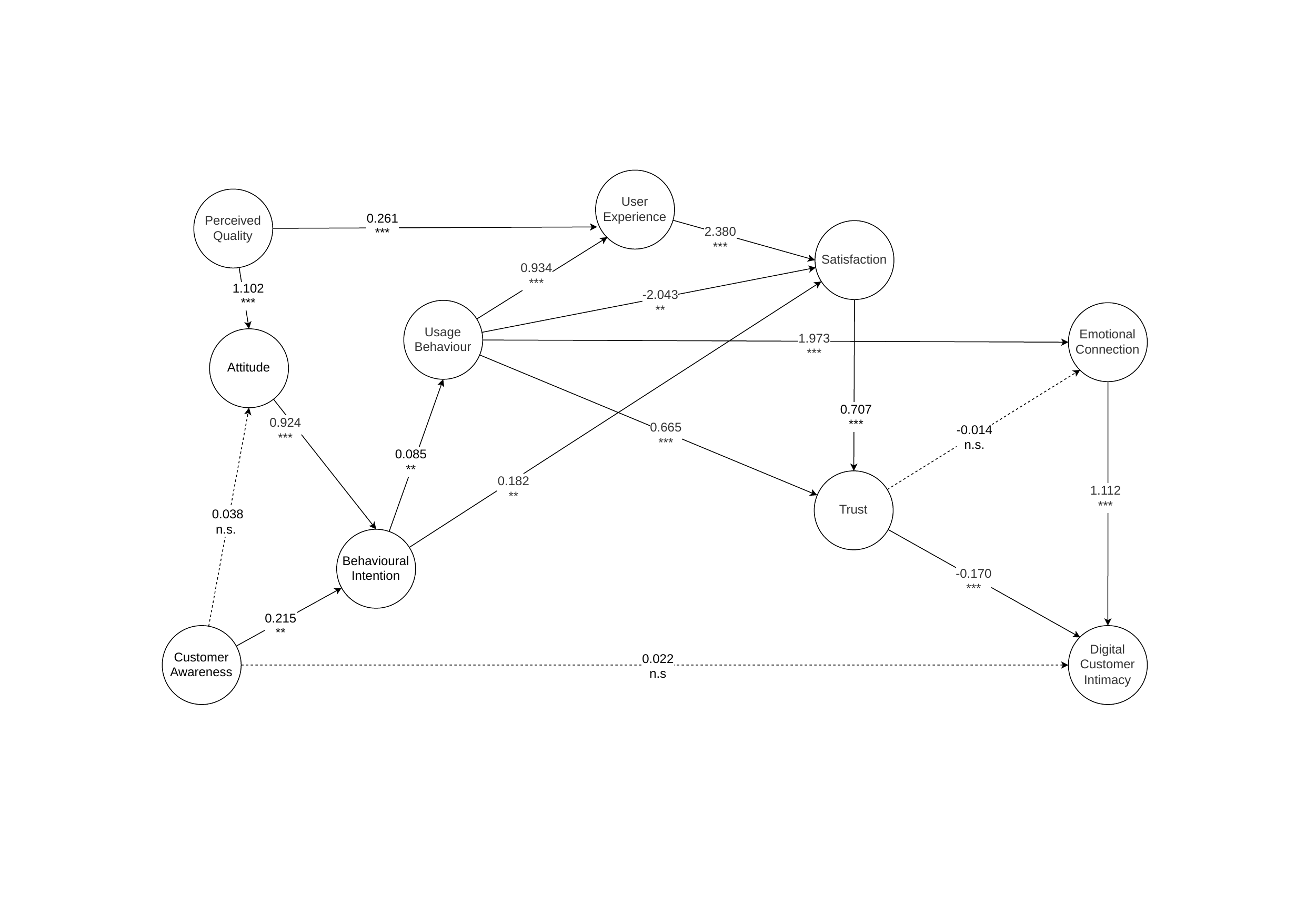} 
\caption{The robustness-baseline SEM structural model used as the basis for the OLS and DML-style robustness stages.}
\label{fig:final_sem_structure}
\end{figure}

\subsection{SEM structural path evidence}
\label{subsec:sem-structural-path-evidence}

Table~\ref{tab:final_sem_structural_paths} reports the 17 tested SEM structural paths. Fourteen paths are statistically significant at the 1\% level or better, while $CA \rightarrow AT$, $TR \rightarrow EC$, and $CA \rightarrow DCI$ are not statistically significant in the SEM model. These non-significant paths are not removed at this stage. Instead, they are carried forward into the OLS, DML, learner-sensitivity, and optional reverse-diagnostic stages so that the final decision can be based on a fuller evidence base.

Table~\ref{tab:sem_r2} reports the explained variance for endogenous constructs in the robustness-baseline SEM model. Most endogenous constructs show substantial to strong explained variance, indicating that the tested structural specification accounts for a large share of variation in the corresponding construct scores. Usage Behaviour has comparatively low explained variance, suggesting that behavioural intention alone explains only a limited portion of actual usage behaviour in this model. The very high explained variance for DCI should be interpreted cautiously, as it may reflect strong shared relational meaning among Digital Customer Intimacy, Emotional Connection, Customer Awareness, and Trust. These $R^2$ values are therefore reported as explanatory diagnostics, not as evidence of causal determination.

\begin{table}[!t]
\centering
\caption{Structural path estimates for the robustness-baseline SEM model using retained indicators}
\label{tab:final_sem_structural_paths}
\compacttable
{\renewcommand{\arraystretch}{1.03}
\begin{tabular}{L{0.22\linewidth}L{0.13\linewidth}L{0.13\linewidth}L{0.13\linewidth}L{0.13\linewidth}L{0.13\linewidth}}
\toprule
\textbf{Path} & \textbf{Estimate} & \textbf{SE} & \textbf{$z$} & \textbf{$p$} & \textbf{Sig.} \\
\midrule
$CA \rightarrow AT$ & 0.038 & 0.062 & 0.608 & 0.543 & n.s. \\
$PQ \rightarrow AT$ & 1.102 & 0.098 & 11.296 & $<0.001$ & *** \\
$AT \rightarrow BI$ & 0.924 & 0.064 & 14.464 & $<0.001$ & *** \\
$CA \rightarrow BI$ & 0.215 & 0.067 & 3.199 & 0.001 & ** \\
$BI \rightarrow UB$ & 0.085 & 0.030 & 2.806 & 0.005 & ** \\
$UB \rightarrow UX$ & 0.934 & 0.261 & 3.580 & $<0.001$ & *** \\
$PQ \rightarrow UX$ & 0.261 & 0.057 & 4.573 & $<0.001$ & *** \\
$BI \rightarrow SA$ & 0.182 & 0.071 & 2.579 & 0.010 & ** \\
$UX \rightarrow SA$ & 2.380 & 0.514 & 4.628 & $<0.001$ & *** \\
$UB \rightarrow SA$ & -2.043 & 0.684 & -2.988 & 0.003 & ** \\
$UB \rightarrow TR$ & 0.665 & 0.198 & 3.366 & $<0.001$ & *** \\
$SA \rightarrow TR$ & 0.707 & 0.049 & 14.287 & $<0.001$ & *** \\
$UB \rightarrow EC$ & 1.973 & 0.530 & 3.723 & $<0.001$ & *** \\
$TR \rightarrow EC$ & -0.014 & 0.095 & -0.147 & 0.883 & n.s. \\
$EC \rightarrow DCI$ & 1.112 & 0.083 & 13.463 & $<0.001$ & *** \\
$TR \rightarrow DCI$ & -0.170 & 0.040 & -4.293 & $<0.001$ & *** \\
$CA \rightarrow DCI$ & 0.022 & 0.040 & 0.551 & 0.582 & n.s. \\
\bottomrule
\end{tabular}
}
\vspace{1mm}
\begin{minipage}{0.96\linewidth}
\footnotesize \textit{Note.} The robustness-baseline SEM model uses the retained/refined measurement indicators after item-level reliability and measurement checks. Non-significant paths are retained at this stage so that path decisions can be informed by SEM, OLS, DML, score-sensitivity, learner-sensitivity, and reverse-diagnostic evidence together.
\end{minipage}
\end{table}

\subsection{Path mapping for OLS and DML-style robustness checks}
\label{subsec:path-mapping-results}

Table~\ref{tab:path_mapping} shows how each tested SEM path is translated into the score-based OLS and DML-style specifications. The selection of the focal predictor and control vector follows the structure of the tested SEM equation. For each SEM-implied path $D \rightarrow Y$, the outcome construct becomes $Y$, the predictor corresponding to the path under examination becomes the focal variable $D$, and the remaining predictors in the same SEM structural equation are included in $\mathbf{X}$ as co-predictor controls. The observed Fintech Type variable is also included as a control in the score-based robustness checks. This design preserves the SEM structural context while allowing each tested path to be examined separately.

\begin{table}[!t]
\centering
\caption{Explained variance for endogenous constructs in the robustness-baseline SEM model}
\label{tab:sem_r2}
\compacttable
\begin{tabular}{L{0.35\linewidth} L{0.18\linewidth} L{0.35\linewidth}}
\toprule
\textbf{Endogenous construct} & \textbf{$R^2$} & \textbf{Interpretation} \\
\midrule
Attitude (AT) & 0.728 & Strong explained variance \\
Behavioural Intention (BI) & 0.827 & Strong explained variance \\
Usage Behaviour (UB) & 0.065 & Low explained variance \\
User Experience (UX) & 0.665 & Substantial explained variance \\
Satisfaction (SA) & 0.881 & Strong explained variance \\
Trust (TR) & 0.784 & Strong explained variance \\
Emotional Connection (EC) & 0.606 & Substantial explained variance \\
Digital Customer Intimacy (DCI) & 0.960 & Very high explained variance \\
\bottomrule
\multicolumn{3}{p{0.90\linewidth}}{\footnotesize \textit{Note.} $R^2$ indicates the proportion of variance in each endogenous construct explained by its predictors in the robustness-baseline SEM model. These values describe explanatory association within the specified model and should not be interpreted as causal evidence.}
\end{tabular}
\end{table}

\begin{table}[!t]
\centering
\caption{Mapping of SEM structural paths to OLS and DML-style specifications}
\label{tab:path_mapping}
\compacttable
{\renewcommand{\arraystretch}{1.03}
\begin{tabular}{L{0.14\linewidth}L{0.22\linewidth}L{0.14\linewidth}L{0.14\linewidth}L{0.26\linewidth}}
\toprule
\textbf{Path ID} & \textbf{SEM path} & \textbf{$Y$} & \textbf{$D$} & \textbf{$\mathbf{X}$} \\
\midrule
TM01 & $CA \rightarrow AT$ & AT & CA & PQ, Fintech Type \\
TM02 & $PQ \rightarrow AT$ & AT & PQ & CA, Fintech Type \\
TM03 & $AT \rightarrow BI$ & BI & AT & CA, Fintech Type \\
TM04 & $CA \rightarrow BI$ & BI & CA & AT, Fintech Type \\
TM05 & $BI \rightarrow UB$ & UB & BI & Fintech Type \\
TM06 & $UB \rightarrow UX$ & UX & UB & PQ, Fintech Type \\
TM07 & $PQ \rightarrow UX$ & UX & PQ & UB, Fintech Type \\
TM08 & $BI \rightarrow SA$ & SA & BI & UX, UB, Fintech Type \\
TM09 & $UX \rightarrow SA$ & SA & UX & BI, UB, Fintech Type \\
TM10 & $UB \rightarrow SA$ & SA & UB & BI, UX, Fintech Type \\
TM11 & $UB \rightarrow TR$ & TR & UB & SA, Fintech Type \\
TM12 & $SA \rightarrow TR$ & TR & SA & UB, Fintech Type \\
TM13 & $UB \rightarrow EC$ & EC & UB & TR, Fintech Type \\
TM14 & $TR \rightarrow EC$ & EC & TR & UB, Fintech Type \\
TM15 & $EC \rightarrow DCI$ & DCI & EC & TR, CA, Fintech Type \\
TM16 & $TR \rightarrow DCI$ & DCI & TR & EC, CA, Fintech Type \\
TM17 & $CA \rightarrow DCI$ & DCI & CA & EC, TR, Fintech Type \\
\bottomrule
\end{tabular}
}
\vspace{1mm}
\begin{minipage}{0.96\linewidth}
\footnotesize \textit{Note.} $Y$ denotes the outcome construct score, $D$ denotes the focal predictor for the SEM-implied path being examined, and $\mathbf{X}$ denotes the control vector used in the OLS and DML-style robustness checks. For paths from a multi-predictor SEM equation, $\mathbf{X}$ contains the remaining predictors in the same SEM equation plus the observed Fintech Type control. For single-predictor equations, $\mathbf{X}$ contains the observed Fintech Type control.
\end{minipage}
\end{table}

\section{Robustness Evaluation}

\subsection{SEM, OLS, and DML comparison using SEM factor scores}
\label{subsec:factor-score-results}

Table~\ref{tab:sem_ols_dml_factor_comparison} reports the SEM, OLS, and DML learner comparison using SEM factor scores. The factor-score results are closely aligned with the SEM measurement model. Most tested paths are supported across OLS and all three DML learners. The weakest path is $CA \rightarrow AT$, which is non-significant in SEM, OLS, and all three DML learners. The paths $TR \rightarrow EC$ and $CA \rightarrow DCI$ are non-significant in SEM but show some score-based robustness evidence, suggesting that they should be reviewed rather than automatically removed.

\begin{table}[!t]
\centering
\caption{SEM, OLS, and DML learner comparison using SEM factor scores}
\label{tab:sem_ols_dml_factor_comparison}
\scriptsize
\setlength{\tabcolsep}{2.4pt}
{\renewcommand{\arraystretch}{1.03}
\begin{adjustbox}{max width=\linewidth}
\begin{tabular}{L{0.15\linewidth}L{0.13\linewidth}L{0.13\linewidth}L{0.15\linewidth}L{0.15\linewidth}L{0.15\linewidth}L{0.09\linewidth}}
\toprule
\textbf{Path} & \textbf{SEM est. (sig.)} & \textbf{OLS est. (sig.)} & \textbf{GBM est. ($p$)} & \textbf{SVM est. ($p$)} & \textbf{RF est. ($p$)} & \textbf{DML sig. learners} \\
\midrule
$CA \rightarrow AT$ & 0.038 (n.s.) & 0.006 (n.s.) & 0.035 (0.171) & 0.025 (0.315) & 0.004 (0.880) & 0/3 \\
$PQ \rightarrow AT$ & 1.102 (***) & 0.915 (***) & 0.894 ($<0.001$) & 0.878 ($<0.001$) & 0.855 ($<0.001$) & 3/3 \\
$AT \rightarrow BI$ & 0.924 (***) & 0.890 (***) & 0.873 ($<0.001$) & 0.872 ($<0.001$) & 0.827 ($<0.001$) & 3/3 \\
$CA \rightarrow BI$ & 0.215 (**) & 0.133 (***) & 0.162 ($<0.001$) & 0.143 ($<0.001$) & 0.132 ($<0.001$) & 3/3 \\
$BI \rightarrow UB$ & 0.085 (**) & 0.363 (***) & 0.355 ($<0.001$) & 0.343 ($<0.001$) & 0.354 ($<0.001$) & 3/3 \\
$UB \rightarrow UX$ & 0.934 (***) & 0.638 (***) & 0.662 ($<0.001$) & 0.654 ($<0.001$) & 0.662 ($<0.001$) & 3/3 \\
$PQ \rightarrow UX$ & 0.261 (***) & 0.417 (***) & 0.384 ($<0.001$) & 0.425 ($<0.001$) & 0.407 ($<0.001$) & 3/3 \\
$BI \rightarrow SA$ & 0.182 (**) & 0.132 (***) & 0.201 ($<0.001$) & 0.209 ($<0.001$) & 0.120 ($<0.001$) & 3/3 \\
$UX \rightarrow SA$ & 2.380 (***) & 1.364 (***) & 1.109 ($<0.001$) & 1.157 ($<0.001$) & 1.244 ($<0.001$) & 3/3 \\
$UB \rightarrow SA$ & -2.043 (**) & -0.782 (***) & -0.567 ($<0.001$) & -0.534 ($<0.001$) & -0.645 ($<0.001$) & 3/3 \\
$UB \rightarrow TR$ & 0.665 (***) & 0.258 (***) & 0.259 ($<0.001$) & 0.279 ($<0.001$) & 0.251 ($<0.001$) & 3/3 \\
$SA \rightarrow TR$ & 0.707 (***) & 0.839 (***) & 0.790 ($<0.001$) & 0.809 ($<0.001$) & 0.765 ($<0.001$) & 3/3 \\
$UB \rightarrow EC$ & 1.973 (***) & 0.995 (***) & 0.975 ($<0.001$) & 0.945 ($<0.001$) & 0.959 ($<0.001$) & 3/3 \\
$TR \rightarrow EC$ & -0.014 (n.s.) & -0.159 (***) & -0.109 ($<0.001$) & -0.120 ($<0.001$) & -0.133 ($<0.001$) & 3/3 \\
$EC \rightarrow DCI$ & 1.112 (***) & 1.044 (***) & 1.008 ($<0.001$) & 0.992 ($<0.001$) & 0.975 ($<0.001$) & 3/3 \\
$TR \rightarrow DCI$ & -0.170 (***) & -0.175 (***) & -0.120 ($<0.001$) & -0.131 ($<0.001$) & -0.137 ($<0.001$) & 3/3 \\
$CA \rightarrow DCI$ & 0.022 (n.s.) & 0.013 (**) & 0.011 (0.456) & 0.023 (0.019) & 0.012 (0.070) & 1/3 \\
\bottomrule
\end{tabular}
\end{adjustbox}
}
\vspace{1mm}
\begin{minipage}{0.96\linewidth}
\footnotesize \textit{Note.} OLS and DML estimates use the same path-level specification as Table~\ref{tab:path_mapping}. DML estimates are derived from the master DML output using Gradient Boosting Machine (GBM), Support Vector Machine regression (SVM), and Random Forest (RF) nuisance learners. The final column reports how many of the three DML learners are statistically significant at $p<0.05$.
\end{minipage}
\end{table}

\subsection{SEM, OLS, and DML comparison using mean composite scores}
\label{subsec:composite-score-results}

Table~\ref{tab:sem_ols_dml_composite_comparison} reports the same comparison using mean composite scores. Most paths remain stable, but several paths become more sensitive. In particular, $UB \rightarrow SA$ is weak under composite-score OLS and all three composite-score DML learners. The $UB \rightarrow TR$ path is supported only by the RF learner under composite scores, while $TR \rightarrow DCI$ is supported by two of the three composite-score DML learners. The $TR \rightarrow EC$ path is especially sensitive because SEM estimates it as non-significant and negative, whereas composite-score OLS and DML estimate it as positive and significant. These patterns show why final path decisions should be made only after comparing SEM, OLS, DML, score-representation, and learner-sensitivity evidence together.

\begin{table}[!t]
\centering
\caption{SEM, OLS, and DML learner comparison using mean composite scores}
\label{tab:sem_ols_dml_composite_comparison}
\scriptsize
\setlength{\tabcolsep}{2.4pt}
{\renewcommand{\arraystretch}{1.03}
\begin{adjustbox}{max width=\linewidth}
\begin{tabular}{L{0.15\linewidth}L{0.13\linewidth}L{0.13\linewidth}L{0.15\linewidth}L{0.15\linewidth}L{0.15\linewidth}L{0.09\linewidth}}
\toprule
\textbf{Path} & \textbf{SEM est. (sig.)} & \textbf{OLS est. (sig.)} & \textbf{GBM est. ($p$)} & \textbf{SVM est. ($p$)} & \textbf{RF est. ($p$)} & \textbf{DML sig. learners} \\
\midrule
$CA \rightarrow AT$ & 0.038 (n.s.) & 0.045 (n.s.) & 0.054 (0.153) & 0.053 (0.168) & 0.036 (0.356) & 0/3 \\
$PQ \rightarrow AT$ & 1.102 (***) & 0.695 (***) & 0.683 ($<0.001$) & 0.670 ($<0.001$) & 0.691 ($<0.001$) & 3/3 \\
$AT \rightarrow BI$ & 0.924 (***) & 0.685 (***) & 0.674 ($<0.001$) & 0.666 ($<0.001$) & 0.665 ($<0.001$) & 3/3 \\
$CA \rightarrow BI$ & 0.215 (**) & 0.158 (***) & 0.173 ($<0.001$) & 0.164 ($<0.001$) & 0.165 ($<0.001$) & 3/3 \\
$BI \rightarrow UB$ & 0.085 (**) & 0.190 (***) & 0.182 ($<0.001$) & 0.172 ($<0.001$) & 0.184 ($<0.001$) & 3/3 \\
$UB \rightarrow UX$ & 0.934 (***) & 0.290 (***) & 0.283 ($<0.001$) & 0.272 ($<0.001$) & 0.282 ($<0.001$) & 3/3 \\
$PQ \rightarrow UX$ & 0.261 (***) & 0.362 (***) & 0.350 ($<0.001$) & 0.359 ($<0.001$) & 0.366 ($<0.001$) & 3/3 \\
$BI \rightarrow SA$ & 0.182 (**) & 0.352 (***) & 0.368 ($<0.001$) & 0.365 ($<0.001$) & 0.357 ($<0.001$) & 3/3 \\
$UX \rightarrow SA$ & 2.380 (***) & 0.358 (***) & 0.396 ($<0.001$) & 0.392 ($<0.001$) & 0.370 ($<0.001$) & 3/3 \\
$UB \rightarrow SA$ & -2.043 (**) & -0.024 (n.s.) & -0.005 (0.920) & 0.004 (0.933) & -0.034 (0.485) & 0/3 \\
$UB \rightarrow TR$ & 0.665 (***) & 0.065 (n.s.) & 0.064 (0.079) & 0.077 (0.059) & 0.075 (0.049) & 1/3 \\
$SA \rightarrow TR$ & 0.707 (***) & 0.685 (***) & 0.679 ($<0.001$) & 0.659 ($<0.001$) & 0.672 ($<0.001$) & 3/3 \\
$UB \rightarrow EC$ & 1.973 (***) & 0.239 (***) & 0.189 ($<0.001$) & 0.213 ($<0.001$) & 0.196 ($<0.001$) & 3/3 \\
$TR \rightarrow EC$ & -0.014 (n.s.) & 0.270 (***) & 0.286 ($<0.001$) & 0.269 ($<0.001$) & 0.289 ($<0.001$) & 3/3 \\
$EC \rightarrow DCI$ & 1.112 (***) & 0.840 (***) & 0.783 ($<0.001$) & 0.775 ($<0.001$) & 0.769 ($<0.001$) & 3/3 \\
$TR \rightarrow DCI$ & -0.170 (***) & -0.098 (***) & -0.057 (0.065) & -0.069 (0.016) & -0.080 (0.009) & 2/3 \\
$CA \rightarrow DCI$ & 0.022 (n.s.) & 0.080 (**) & 0.091 (0.004) & 0.092 (0.001) & 0.094 ($<0.001$) & 3/3 \\
\bottomrule
\end{tabular}
\end{adjustbox}
}
\vspace{1mm}
\begin{minipage}{0.96\linewidth}
\footnotesize \textit{Note.} OLS and DML estimates use the same path-level specification as Table~\ref{tab:path_mapping}. DML estimates are derived from the master DML output using Gradient Boosting Machine (GBM), Support Vector Machine regression (SVM), and Random Forest (RF) nuisance learners. The final column reports how many of the three DML learners are statistically significant at $p<0.05$.
\end{minipage}
\end{table}

\subsection{Reverse-direction diagnostic results}
\label{subsec:reverse-direction-results}

Table~\ref{tab:reverse_direction_diagnostics} summarises the reverse-direction diagnostics. The diagnostics include core path pairs that are conceptually important for the DCI model and optional path pairs added because the tested SEM path was non-significant or because the relationship is useful for model-decision review. These checks are interpreted only as directional association diagnostics. They do not establish causal direction.

\begin{table}[!t]
\centering
\caption{Reverse-direction diagnostic checks for selected path pairs}
\label{tab:reverse_direction_diagnostics}
\scriptsize
\setlength{\tabcolsep}{2.4pt}
{\renewcommand{\arraystretch}{1.05}
\begin{adjustbox}{max width=\linewidth}
\begin{tabular}{L{0.15\linewidth}L{0.14\linewidth}L{0.27\linewidth}L{0.27\linewidth}L{0.22\linewidth}}
\toprule
\textbf{Pair} & \textbf{Status} & \textbf{Factor-score support} & \textbf{Composite-score support} & \textbf{Diagnostic interpretation} \\
\midrule
$CA \leftrightarrow AT$ & optional diagnostic & $CA \rightarrow AT$: 0/3; $AT \rightarrow CA$: 0/3 & $CA \rightarrow AT$: 0/3; $AT \rightarrow CA$: 0/3 & Weak in both directions; supports end-stage review. \\
$CA \leftrightarrow DCI$ & optional diagnostic & $CA \rightarrow DCI$: 1/3; $DCI \rightarrow CA$: 1/3 & $CA \rightarrow DCI$: 3/3; $DCI \rightarrow CA$: 3/3 & Score-sensitive; composite scores support both directions. \\
$UB \leftrightarrow SA$ & core diagnostic & $UB \rightarrow SA$: 3/3; $SA \rightarrow UB$: 3/3 & $UB \rightarrow SA$: 0/3; $SA \rightarrow UB$: 0/3 & Factor-score support but composite-score weakness; score-sensitive. \\
$TR \leftrightarrow DCI$ & core diagnostic & $TR \rightarrow DCI$: 3/3; $DCI \rightarrow TR$: 3/3 & $TR \rightarrow DCI$: 2/3; $DCI \rightarrow TR$: 3/3 & Reverse direction is stable; indicates directional complexity. \\
$UB \leftrightarrow TR$ & optional diagnostic & $UB \rightarrow TR$: 3/3; $TR \rightarrow UB$: 3/3 & $UB \rightarrow TR$: 1/3; $TR \rightarrow UB$: 1/3 & Factor-score support but weak composite support; score-sensitive. \\
$EC \leftrightarrow DCI$ & core diagnostic & $EC \rightarrow DCI$: 3/3; $DCI \rightarrow EC$: 3/3 & $EC \rightarrow DCI$: 3/3; $DCI \rightarrow EC$: 3/3 & Stable in both directions; possible reciprocal association. \\
$TR \leftrightarrow EC$ & optional diagnostic & $TR \rightarrow EC$: 1/3; $EC \rightarrow TR$: 1/3 & $TR \rightarrow EC$: 3/3; $EC \rightarrow TR$: 3/3 & Score-sensitive and sign-sensitive; review theory. \\
\bottomrule
\end{tabular}
\end{adjustbox}
}
\vspace{1mm}
\begin{minipage}{0.96\linewidth}
\footnotesize \textit{Note.} Entries such as 3/3 indicate that all three DML learners were significant for that direction. Reverse-direction diagnostics are reported as association-level sensitivity checks only. They do not establish reverse or bidirectional causality.
\end{minipage}
\end{table}

The reverse diagnostics add several useful insights. First, $EC \leftrightarrow DCI$ is stable in both directions across both score types, suggesting a strong reciprocal association pattern between emotional connection and digital customer intimacy. Second, $TR \leftrightarrow DCI$ is directionally complex: the reverse diagnostic $DCI \rightarrow TR$ is stable, while $TR \rightarrow DCI$ is weaker under composite scoring. Third, $UB \leftrightarrow SA$ and $UB \leftrightarrow TR$ are strongly supported under factor scores but weak under composite scores, reinforcing the conclusion that usage-related relational paths are score-sensitive. Fourth, the optional diagnostics for $CA \leftrightarrow AT$, $CA \leftrightarrow DCI$, and $TR \leftrightarrow EC$ provide additional evidence for end-stage model review rather than automatic path removal.

\subsection{Integrated model-decision evidence}
\label{subsec:integrated-model-decision-evidence}

Table~\ref{tab:integrated_robustness_classification} integrates the SEM, OLS, DML, score-sensitivity, learner-sensitivity, and reverse-diagnostic evidence. The table should be read as decision support rather than as a mechanical rule for retaining or removing paths. The strongest candidates for retention are paths that are significant in SEM and stable across OLS, factor-score DML, composite-score DML, and learners. Paths with weak SEM support and weak downstream support, such as $CA \rightarrow AT$, are candidates for removal, revision, or theoretical reconsideration during final structural path evaluation. Paths with mixed SEM and score-based evidence, such as $TR \rightarrow EC$ and $CA \rightarrow DCI$, require theory and measurement review before a final decision is made.

\begin{table}[!t]
\centering
\caption{Integrated evidence summary for model-decision support}
\label{tab:integrated_robustness_classification}
\scriptsize
\setlength{\tabcolsep}{2.4pt}
{\renewcommand{\arraystretch}{1.05}
\begin{adjustbox}{max width=\linewidth}
\begin{tabular}{L{0.13\linewidth}L{0.08\linewidth}L{0.20\linewidth}L{0.18\linewidth}L{0.20\linewidth}L{0.22\linewidth}}
\toprule
\textbf{Path} & \textbf{SEM sig.} & \textbf{Factor-score evidence} & \textbf{Composite-score evidence} & \textbf{Evidence label} & \textbf{Decision-support use} \\
\midrule
$CA \rightarrow AT$ & n.s. & OLS n.s.; RF n.s.; DML partial/no & direction Yes; DML partial/no & Weak empirical support & Candidate trim/review if theory is weak \\
$PQ \rightarrow AT$ & *** & OLS ***; RF ***; DML all & direction Yes; DML all & Strong retention evidence & Candidate retain \\
$AT \rightarrow BI$ & *** & OLS ***; RF ***; DML all & direction Yes; DML all & Strong retention evidence & Candidate retain \\
$CA \rightarrow BI$ & ** & OLS ***; RF ***; DML all & direction Yes; DML all & Strong retention evidence & Candidate retain \\
$BI \rightarrow UB$ & ** & OLS ***; RF ***; DML all & direction Yes; DML all & Strong retention evidence & Candidate retain \\
$UB \rightarrow UX$ & *** & OLS ***; RF ***; DML all & direction Yes; DML all & Strong retention evidence & Candidate retain \\
$PQ \rightarrow UX$ & *** & OLS ***; RF ***; DML all & direction Yes; DML all & Strong retention evidence & Candidate retain \\
$BI \rightarrow SA$ & ** & OLS ***; RF ***; DML all & direction Yes; DML all & Strong retention evidence & Candidate retain \\
$UX \rightarrow SA$ & *** & OLS ***; RF ***; DML all & direction Yes; DML all & Strong retention evidence & Candidate retain \\
$UB \rightarrow SA$ & ** & OLS ***; RF ***; DML all & direction Yes; DML partial/no & Strong retention evidence & Candidate retain, but discuss sensitivity \\
$UB \rightarrow TR$ & *** & OLS ***; RF ***; DML all & direction Yes; DML partial/no & Strong retention evidence & Candidate retain, but discuss sensitivity \\
$SA \rightarrow TR$ & *** & OLS ***; RF ***; DML all & direction Yes; DML all & Strong retention evidence & Candidate retain \\
$UB \rightarrow EC$ & *** & OLS ***; RF ***; DML all & direction Yes; DML all & Strong retention evidence & Candidate retain \\
$TR \rightarrow EC$ & n.s. & OLS ***; RF ***; DML all & direction Check; DML all & Mixed evidence; theory review needed & Theory and measurement review before decision \\
$EC \rightarrow DCI$ & *** & OLS ***; RF ***; DML all & direction Yes; DML all & Strong retention evidence & Candidate retain \\
$TR \rightarrow DCI$ & *** & OLS ***; RF ***; DML all & direction Yes; DML partial/no & Strong retention evidence & Candidate retain, but discuss sensitivity \\
$CA \rightarrow DCI$ & n.s. & OLS **; RF n.s.; DML partial/no & direction Yes; DML all & Mixed evidence; theory review needed & Theory and measurement review before decision \\
\bottomrule
\end{tabular}
\end{adjustbox}
}
\vspace{1mm}
\begin{minipage}{0.96\linewidth}
\footnotesize \textit{Note.} This table provides model-decision support, not automatic path-decision rules. Final decisions should consider theory, construct validity, SEM fit, parsimony, robustness evidence, and substantive interpretability together.
\end{minipage}
\end{table}

Table~\ref{tab:rq_answers} summarises the answers to the main research question, sub-research questions, and applied demonstration question.

\begin{table}[!t]
\centering
\caption{Summary answers to the research questions}
\label{tab:rq_answers}
\compacttable
{\renewcommand{\arraystretch}{1.18}
\begin{tabular}{L{0.10\linewidth} L{0.16\linewidth} L{0.66\linewidth}}
\toprule
\textbf{Question} & \textbf{Focus} & \textbf{Summary answer} \\
\midrule

RQ1
&
Methodological framework
&
SEM, OLS, and DML-style analysis can be integrated into a staged robustness framework that separates measurement refinement from final structural path evaluation. SEM provides the latent-variable measurement and theory-testing foundation, OLS provides a transparent score-based regression benchmark, and DML-style residualisation provides flexible path-level robustness evidence. The framework allows researchers to evaluate an initially tested SEM model before making final structural model refinement decisions. \\

RQ2
&
SEM-to-regression translation
&
A robustness-baseline SEM model can be translated into OLS and DML-style specifications by first constructing respondent-level construct scores from the refined measurement model. Each initially tested SEM path is then represented as a path-level specification with an outcome construct $Y$, a focal predictor construct $D$, and a control vector $\mathbf{X}$. For multi-predictor SEM equations, the focal predictor is placed in $D$, while the remaining same-equation predictors are included in $\mathbf{X}$ as co-predictor controls. \\

RQ3
&
Complementary robustness evidence
&
OLS and DML-style estimates complement SEM by examining whether initially tested structural paths remain visible under alternative estimation assumptions. OLS provides a simple and interpretable linear benchmark using construct scores, while DML-style analysis flexibly adjusts for SEM co-predictors and observed controls. Together, they help distinguish paths that are stable across methods from paths that are sensitive to score construction, model specification, or flexible-control adjustment. \\

RQ4
&
Robustness interpretation
&
Convergent evidence across SEM, OLS, DML-style estimates, score representations, and learners can be interpreted as stronger robustness evidence. Divergent evidence should be treated diagnostically: score-type differences indicate measurement-representation sensitivity, learner differences indicate nuisance-learner sensitivity, and selected reverse-direction diagnostics indicate possible reciprocal association or directional instability. These diagnostics support interpretation and final structural path evaluation, but they do not establish causal direction in cross-sectional survey data. \\

RQ5
&
DCI demonstration case
&
In the FinTech DCI demonstration, most initially tested structural relationships remain stable across the SEM--OLS--DML robustness workflow. The framework also identifies paths requiring more cautious interpretation or further review, including weakly supported paths such as $CA \rightarrow AT$, mixed-evidence paths such as $TR \rightarrow EC$ and $CA \rightarrow DCI$, and directionally or score-sensitive paths such as $UB \rightarrow SA$, $UB \rightarrow TR$, and $TR \rightarrow DCI$. These findings illustrate how the proposed framework can support final structural model refinement decisions after multiple sources of empirical and theoretical evidence are considered together. \\

\bottomrule
\end{tabular}
}
\end{table}

\section{Discussion}

\subsection{Methodological implications}

The workflow demonstrates a more conservative and transparent way to use SEM--OLS--DML robustness analysis. Rather than removing structural paths immediately after SEM estimation and then checking only the surviving paths, our implementation tests the full theory-driven SEM model before final structural path evaluation. This allows the researcher to review SEM significance, OLS benchmarks, DML-style estimates, learner sensitivity, score-representation sensitivity, and reverse-direction diagnostics together before making final structural decisions.

This shift strengthens the methodological contribution. SEM remains central because it validates the measurement model and estimates the theory-specified latent path system. OLS and DML-style checks are not replacements for SEM. They are supplementary evidence layers that help identify whether a path is strongly supported, score-sensitive, learner-sensitive, directionally complex, or empirically weak. The resulting workflow is therefore closer to a model-decision support process than a simple robustness appendix.

\subsection{How the approach differs from typical DML and SEM--ML use}

The proposed approach differs from typical applied DML studies because DML is not used here as the primary estimator for a single observed treatment--outcome relationship. Instead, SEM first provides the measurement model and tested path system. DML is then applied path by path to examine whether the SEM-implied associations remain stable after flexible adjustment for observed controls and co-predictors. This adaptation is especially relevant for survey-based research because direct application of DML to raw items or simple composite scores can underuse SEM's ability to represent latent constructs and measurement quality.

The approach also differs from SEM--ML integration work that uses machine learning mainly for prediction, nonlinear discovery, interaction detection, or alternative model exploration. The purpose here is diagnostic robustness and model-decision support. The central question is not whether machine learning can outperform SEM predictively, but whether SEM-implied associations remain directionally and statistically stable under transparent regression, flexible-control residualisation, alternative score construction, and alternative nuisance learners.

\subsection{Substantive implications for the FinTech DCI case}

The demonstration suggests that much of the adoption-to-intimacy pathway is stable. Perceived quality, attitude, customer awareness, behavioural intention, usage behaviour, user experience, satisfaction, trust, emotional connection, and DCI are connected in a broadly coherent sequence. The strongest downstream relational evidence concerns emotional connection and DCI: $EC \rightarrow DCI$ remains strongly supported across score types and learners, and the reverse-direction diagnostic also suggests that emotional connection and digital customer intimacy may be mutually reinforcing. This is consistent with customer-experience thinking that treats emotional connection as a deeper relational outcome than satisfaction alone \citep{zorfas2016emotionalconnection}.

The usage-related paths require more nuanced interpretation. The $UB \rightarrow SA$ path is negative and significant in SEM and strongly supported under factor-score OLS and DML, but it becomes weak under composite-score OLS and DML. This indicates score-representation sensitivity rather than a simple conclusion that usage reduces satisfaction. Prior IS research has treated system usage behaviour as closely related to user satisfaction and even as a potential proxy for satisfaction measurement \citep{downing1999usage}; therefore, the present result should be discussed as a conditional association that may depend on how usage behaviour and satisfaction are operationalised.

The usage--trust relationship also requires cautious interpretation. In the tested SEM specification, the model explicitly places usage behaviour as the predictor and trust as the outcome; that is, $UB \rightarrow TR$ treats usage behaviour as an antecedent of trust. The positive and significant SEM estimate indicates that higher usage behaviour is associated with higher trust, conditional on the model specification. This path is strongly supported in the factor-score robustness analysis, but the composite-score evidence is weaker, suggesting some sensitivity to how the constructs are represented. Prior mobile-application and FinTech studies also support a close relationship between trust and use behaviour, but they often specify the relationship in the opposite direction, with trust acting as an antecedent of behavioural intention or actual use \citep{yan2013mobiletrust,ratnawati2022flip}. Therefore, the $UB \leftrightarrow TR$ relationship is better interpreted as a directionally sensitive association. Rather than concluding that usage behaviour unidirectionally determines trust, the evidence suggests that usage behaviour and trust may be mutually associated and should be examined further in future longitudinal or process-oriented research.

The trust-to-intimacy and trust-to-emotional-connection results further illustrate the value of testing before final structural path evaluation. The $TR \rightarrow DCI$ path is negative in SEM and remains negative under most robustness checks, but the reverse diagnostic $DCI \rightarrow TR$ is also stable. This suggests directional complexity, possible suppression by emotional connection, or a distinction between functional trust and deeper intimacy. The $TR \rightarrow EC$ path is non-significant in SEM but becomes significant in score-based robustness checks, with sign sensitivity between factor and composite scores. This path should therefore be reviewed theoretically and empirically before a final model decision is made.

Finally, the $CA \rightarrow AT$ and $CA \rightarrow DCI$ paths show why the revised workflow is useful. $CA \rightarrow AT$ is weak in SEM and weak across downstream checks, making it a candidate for removal or theoretical reconsideration if theory is also weak. By contrast, $CA \rightarrow DCI$ is non-significant in SEM but shows stronger composite-score robustness evidence, suggesting that it should be reviewed rather than automatically removed. These examples demonstrate that path decisions should not be made from SEM $p$-values alone.

\subsection{Boundary conditions and limitations}

Several limitations should be recognised. First, DML-style residualisation adjusts flexibly for observed controls, but it does not remove unobserved confounding by itself. Second, the DML-style models used here retain a partially linear focal path: machine learning is used to model the nuisance control functions, not to claim a fully nonlinear causal effect of the focal predictor. Third, factor-score results are closely aligned with the SEM measurement model and should be interpreted alongside composite-score sensitivity checks. Fourth, very high $R^2$ values in score-based downstream models should be examined carefully because they may reflect the construction and scaling of SEM factor scores rather than purely substantive explanatory power. Fifth, reverse-direction checks should be interpreted only as directional association diagnostics, not as evidence of bidirectional causality. Longitudinal, experimental, quasi-experimental, or process-modelling designs would be needed to make stronger claims about temporal ordering or causal direction.

\section{Conclusion}

This paper proposes and demonstrates a staged SEM--OLS--DML robustness framework for survey-based latent-construct research. The proposed workflow estimates the robustness-baseline SEM model before final structural path evaluation, then uses OLS, DML-style residualisation, score-representation sensitivity, learner sensitivity, and selected reverse-direction diagnostics to generate model-decision evidence. In the FinTech Digital Customer Intimacy demonstration, most tested paths remain stable across methods and score types. The strongest downstream relational path remains $EC \rightarrow DCI$. The revised evidence also identifies paths requiring end-stage review: $CA \rightarrow AT$ shows weak empirical support, $TR \rightarrow EC$ and $CA \rightarrow DCI$ show mixed SEM and score-based evidence, and $UB \rightarrow SA$, $UB \rightarrow TR$, and $TR \rightarrow DCI$ require cautious discussion because of score sensitivity or directional complexity. The framework provides a practical interpretation guide for researchers seeking to complement SEM with conventional and machine-learning-based robustness checks. The accompanying Colab workbook and Zenodo archive further position the paper as a reusable template for researchers and students who wish to adapt the workflow to other survey-based SEM applications.

\section*{Data and Code Availability}

The Google Colab workbook, generated result tables, and selected output figures associated with this study are available from Zenodo (\url{https://doi.org/10.5281/zenodo.21073457}). The archive is intended to serve both as a replication package for the empirical demonstration and as a reusable template for applying the SEM--OLS--DML robustness workflow to other survey-based latent-construct datasets.


\begin{thebibliography}{00}

\bibitem[Anderson and Gerbing(1988)]{anderson1988sem}
Anderson, J. C., and Gerbing, D. W. (1988).
Structural equation modeling in practice: A review and recommended two-step approach.
\textit{Psychological Bulletin, 103}(3), 411--423.

\bibitem[Bach et al.(2022)]{bach2022doubleml}
Bach, P., Chernozhukov, V., Kurz, M. S., and Spindler, M. (2022).
DoubleML---An object-oriented implementation of double machine learning in Python.
\textit{Journal of Machine Learning Research, 23}(53), 1--6.

\bibitem[Bareinboim et al.(2022)]{bareinboim2022pearlhierarchy}
Bareinboim, E., Correa, J. D., Ibeling, D., and Icard, T. (2022).
On Pearl's hierarchy and the foundations of causal inference.
In H. Geffner, R. Dechter, and J. Y. Halpern (Eds.),
\textit{Probabilistic and causal inference: The works of Judea Pearl}
(pp. 507--556).
ACM Books.

\bibitem[Bhattacherjee(2001)]{bhattacherjee2001continuance}
Bhattacherjee, A. (2001).
Understanding information systems continuance: An expectation-confirmation model.
\textit{MIS Quarterly, 25}(3), 351--370.

\bibitem[Bollen(1989)]{bollen1989sem}
Bollen, K. A. (1989).
\textit{Structural equations with latent variables}.
John Wiley \& Sons.

\bibitem[Breiman(2001)]{breiman2001randomforests}
Breiman, L. (2001).
Random forests.
\textit{Machine Learning, 45}(1), 5--32.

\bibitem[Brunner(2023)]{brunner2023openSEM}
Brunner, J. (2023).
\textit{Structural equation models: An open textbook}
(Edition 0.10).
Department of Statistical Sciences, University of Toronto.

\bibitem[Chernozhukov et al.(2018)]{chernozhukov2018dml}
Chernozhukov, V., Chetverikov, D., Demirer, M., Duflo, E., Hansen, C., Newey, W., and Robins, J. (2018).
Double/debiased machine learning for treatment and structural parameters.
\textit{The Econometrics Journal, 21}(1), C1--C68.

\bibitem[Cortes and Vapnik(1995)]{cortes1995svm}
Cortes, C., and Vapnik, V. (1995).
Support-vector networks.
\textit{Machine Learning, 20}(3), 273--297.

\bibitem[Davis(1989)]{davis1989tam}
Davis, F. D. (1989).
Perceived usefulness, perceived ease of use, and user acceptance of information technology.
\textit{MIS Quarterly, 13}(3), 319--340.


\bibitem[Downing(1999)]{downing1999usage}
Downing, C. E. (1999).
System usage behavior as a proxy for user satisfaction: An empirical investigation.
\textit{Information \& Management, 35}(4), 203--216.
\url{https://doi.org/10.1016/S0378-7206(98)00090-1}.

\bibitem[Friedman(2001)]{friedman2001gbm}
Friedman, J. H. (2001).
Greedy function approximation: A gradient boosting machine.
\textit{The Annals of Statistics, 29}(5), 1189--1232.

\bibitem[Gefen et al.(2000)]{gefen2000semregression}
Gefen, D., Straub, D. W., and Boudreau, M.-C. (2000).
Structural equation modeling and regression: Guidelines for research practice.
\textit{Communications of the Association for Information Systems, 4},
Article 7.

\bibitem[Liu et al.(2024a)]{liu2024fintechmapping}
Liu, Q., Chan, K.-C., and Chimhundu, R. (2024a).
Fintech research: Systematic mapping, classification, and future directions.
\textit{Financial Innovation, 10}(1), Article 24.

\bibitem[Liu et al.(2024b)]{liu2024digitalcustomerintimacy}
Liu, Q., Chan, K.-C., and Chimhundu, R. (2024b).
From customer intimacy to digital customer intimacy.
\textit{Journal of Theoretical and Applied Electronic Commerce Research, 19}(4), 3386--3411.

\bibitem[Liu et al.(2026)]{liu2026dcimodel}
Liu, Q., Chan, K.-C., Tiwari, S., and Chimhundu, R. (2026).
\textit{From adoption to intimacy: Experiential and emotional pathways of digital customer intimacy in FinTech}.
Manuscript under review.

\bibitem[MacKenzie et al.(2011)]{mackenzie2011construct}
MacKenzie, S. B., Podsakoff, P. M., and Podsakoff, N. P. (2011).
Construct measurement and validation procedures in MIS and behavioral research: Integrating new and existing techniques.
\textit{MIS Quarterly, 35}(2), 293--334.

\bibitem[Pearl(2009)]{pearl2009causality}
Pearl, J. (2009).
\textit{Causality: Models, reasoning, and inference}
(2nd ed.).
Cambridge University Press.

\bibitem[Podsakoff et al.(2003)]{podsakoff2003cmb}
Podsakoff, P. M., MacKenzie, S. B., Lee, J.-Y., and Podsakoff, N. P. (2003).
Common method biases in behavioral research: A critical review of the literature and recommended remedies.
\textit{Journal of Applied Psychology, 88}(5), 879--903.


\bibitem[Ratnawati et al.(2022)]{ratnawati2022flip}
Ratnawati, S., Durachman, Y., and Saputra, A. (2022).
Analyzing factors influencing intention to use and actual use of mobile fintech applications free interbank money transfer Flip using UTAUT 2 model with trust and perceived security.
In \textit{2022 10th International Conference on Cyber and IT Service Management (CITSM)}.
\url{https://doi.org/10.1109/CITSM56380.2022.9935838}.

\bibitem[Richter and Tudoran(2024)]{richter2024plsml}
Richter, N. F., and Tudoran, A. A. (2024).
Elevating theoretical insight and predictive accuracy in business research: Combining PLS-SEM and selected machine learning algorithms.
\textit{Journal of Business Research, 173}, Article 114453.

\bibitem[Shi et al.(2025)]{shi2025dmlguide}
Shi, B., Mao, X., Yang, M., and Li, B. (2025).
What, why, and how: An empiricist's guide to double/debiased machine learning.
\textit{Information Systems Research}.
Advance online publication.

\bibitem[Treacy and Wiersema(1993)]{treacy1993customerintimacy}
Treacy, M., and Wiersema, F. (1993).
Customer intimacy and other value disciplines.
\textit{Harvard Business Review, 71}(1), 84--93.

\bibitem[Venkatesh et al.(2003)]{venkatesh2003utaut}
Venkatesh, V., Morris, M. G., Davis, G. B., and Davis, F. D. (2003).
User acceptance of information technology: Toward a unified view.
\textit{MIS Quarterly, 27}(3), 425--478.

\bibitem[Wu et al.(2024)]{wu2024fintechgreen}
Wu, B., Ding, Y., Xie, B., and Zhang, Y. (2024).
FinTech and inclusive green growth: A causal inference based on double machine learning.
\textit{Sustainability, 16}(22), Article 9989.
\url{https://doi.org/10.3390/su16229989}.

\bibitem[Yan et al.(2013)]{yan2013mobiletrust}
Yan, Z., Dong, Y., Niemi, V., and Yu, G. (2013).
Exploring trust of mobile applications based on user behaviors: An empirical study.
\textit{Journal of Applied Social Psychology, 43}(3), 638--659.
\url{https://doi.org/10.1111/j.1559-1816.2013.01044.x}.


\bibitem[Zorfas and Leemon(2016)]{zorfas2016emotionalconnection}
Zorfas, A., and Leemon, D. (2016, August 29).
An emotional connection matters more than customer satisfaction.
\textit{Harvard Business Review}.
\url{https://hbr.org/2016/08/an-emotional-connection-matters-more-than-customer-satisfaction}.

\end{thebibliography}
\end{document}